\documentclass[runningheads]{llncs}
\usepackage[T1]{fontenc}

\usepackage[pagebackref,breaklinks,colorlinks]{hyperref}
\usepackage{amsfonts}
\usepackage{algorithmic}
\usepackage{graphicx}
\usepackage{textcomp}
\usepackage{booktabs}
\usepackage{multirow}
\usepackage[table,xcdraw]{xcolor}
\usepackage{makecell}
\usepackage{multirow}
\usepackage{subcaption}
\usepackage{lipsum}

\usepackage{times,epsf,mathtools,xcolor,amssymb,amsmath,enumitem,wrapfig,comment,cleveref,appendix,float}

\newcommand{\PD}{\mathrm{PD}}

\newcommand{\X}{\mathcal{X}}

\newcommand{\Y}{\mathbf{Y}}

\begin{document}
\title{TopOC: Topological Deep Learning\\ for Ovarian and Breast Cancer Diagnosis}
\author{Saba Fatema\inst{1} \and
Brighton Nuwagira\inst{1}\and
Sayoni Chakraborty\inst{1}\and \\
Reyhan Gedik\inst{2}\and
Baris Coskunuzer\inst{1}}
%\orcidID{0000-0001-7462-8819}
\authorrunning{S. Fatema et al.}
\institute{U. Texas at Dallas, Dept. Math. Sciences, Richardson, TX 75080 USA \and
Harvard Medical School, MGH - Pathology Dept., Boston, MA 02114 USA\\
\email{\{saba.fatema,brighton.nuwagira,sayoni.chakraborty, coskunuz\}@utdallas.edu, rgedik@mgh.harvard.edu}}
\maketitle              % typeset the header of the contribution
\vspace{-.2in}
\begin{abstract}
Microscopic examination of slides prepared from tissue samples is the primary tool for detecting and classifying cancerous lesions, a process that is time-consuming and requires the expertise of experienced pathologists. Recent advances in deep learning methods hold significant potential to enhance medical diagnostics and treatment planning by improving accuracy, reproducibility, and speed, thereby reducing clinicians' workloads and turnaround times. However, the necessity for vast amounts of labeled data to train these models remains a major obstacle to the development of effective clinical decision support systems.

In this paper, we propose the integration of topological deep learning methods to enhance the accuracy and robustness of existing histopathological image analysis models. Topological data analysis (TDA) offers a unique approach by extracting essential information through the evaluation of topological patterns across different color channels. While deep learning methods capture local information from images, TDA features provide complementary global features. Our experiments on publicly available histopathological datasets demonstrate that the inclusion of topological features significantly improves the differentiation of tumor types in ovarian and breast cancers.

\keywords{Ovarian Cancer Diagnosis \and Breast Cancer Diagnosis \and Histopathology \and Cubical Persistence \and Topological Data Analysis}
\end{abstract}

\section{Introduction} \label{sec:intro}
%\BC{Reyhan, please revise the intro as you like. It would be great if you could add some domain-specific details for these cancer types.}

Ovarian and breast cancers rank among the most common and lethal malignancies impacting women globally. The cornerstone of detecting and classifying these cancers is histopathology, which involves the microscopic examination of tissue samples and cellular collections. This meticulous process is vital for accurate diagnosis and treatment planning but is often time-consuming and heavily dependent on the expertise of seasoned pathologists. As the demand for precise and rapid diagnostics increases, the constraints of traditional methods become more evident, underscoring the need for advanced technologies to support and enhance pathological workflows.

Recent advancements in deep learning have shown great promise in the field of medical imaging, offering the potential to revolutionize diagnostics by providing improved accuracy, reproducibility, and efficiency. Deep learning algorithms excel at identifying complex patterns within images, making them well-suited for the detailed analysis required in histopathology. However, the effectiveness of these models is often hindered by the substantial requirement for large, labeled datasets, which can be difficult and costly to obtain.

To address these challenges, we explore the integration of topological data analysis (TDA) tools with deep learning methods for histopathological image analysis. TDA provides a novel approach by focusing on the extraction and evaluation of topological patterns within images, capturing global features that complement the local information typically identified by deep learning algorithms. This synergy between local and global feature extraction can enhance the robustness and accuracy of cancer detection models.

In this paper, we present a method that incorporates topological deep learning techniques into the analysis of histopathological images. We demonstrate the effectiveness of this approach using publicly available benchmark datasets for ovarian and breast cancer, showing that our method significantly improves the differentiation of tumor types. Our results suggest that the integration of TDA with deep learning not only enhances model performance but also has the potential to streamline diagnostic workflows, ultimately benefiting both clinicians and patients.

Our contributions can be summarized as follows:

\noindent $\diamond$  We introduce two robust topological machine learning models for the analysis of histopathological images.

\noindent $\diamond$  Using cubical persistence, we extract unique topological fingerprints by examining the evolution of topological patterns in the images across each color channel.

\noindent $\diamond$  Our first model, TopOC-1, applies standard machine learning models to topological vectors, making it straightforward, interpretable, and computationally feasible without the need for data augmentation.

\noindent $\diamond$ Our second model, TopOC-CNN, integrates topological features with existing pre-trained models. We observe consistent performance improvements with the inclusion of these topological features.

\noindent $\diamond$  Both models deliver outstanding performance on ovarian and breast cancer histopathological image datasets, surpassing state-of-the-art models.

\vspace{-.1in}

\section{Related Work} \label{sec:related}

\subsection{Machine Learning Methods for Ovarian and Breast Cancer Diagnosis}

%\BC{Saba, please add a paragraph for breast cancer methods.}

% \BC{Saba, write 2-3 paragraph overview of ML in OCD with emphasis on deep learning. Use~\cite{liberto2022current,sadeghi2024deep} or similar surveys}

Deep learning (DL) techniques show significant potential in enhancing ovarian cancer diagnosis through histopathological image analysis, identifying malignant patterns, and aiding clinicians in decision-making~\cite{sadeghi2024deep}. \cite{farahani2022deep} presents DL model for ovarian carcinoma histotype classification comparable to expert pathologists. Various CNN models like YOLO, DenseNet, GoogleNet(V3), ResNet, etc., have demonstrated competitive accuracy with medical professionals using ultrasound images \cite{wu2018deep,pham2024ovarian}.

A hybrid DL model using multi-modal data (gene and histopathology images) is proposed by \cite{ghoniem2021multi} to address the limitations of single-modal approaches in representing ovarian cancer characteristics. An approach for better ovarian cancer risk stratification by integrating histopathological, radiologic, and clinicogenomic data is demonstrated in \cite{boehm2022multimodal}. A review in \cite{breen2023artificial} assesses AI in ovarian cancer pathology, identifies clinical gaps, and recommends improvements for future adoption.

DL methods are also employed in various recent studies to detect breast cancer. In \cite{khalid2023breast}, authors introduce a model for detecting breast cancer in mammograms of varying densities, using low-variance elimination, univariate selection, and recursive feature elimination as feature selection modules. \cite{yadav2023diagnosis} proposed a multimodal machine learning model for breast cancer detection that combines mammograms and ultrasound data. For a thorough review of the breast cancer histopathological image analysis methods, see \cite{zhou2020comprehensive,chan2023artificial}, where the authors review the recent models based on artificial neural networks, categorizing them into classical and deep neural network approaches.

\vspace{-.1in}

\subsection{TDA in Medical Image Analysis}

Persistent homology (PH) has been highly effective for pattern recognition in image and shape analysis over the past two decades. In medical image analysis, PH has produced powerful results in cell development~\cite{mcguirl2020topological}, tumor detection~\cite{crawford2020predicting,wang2021topotxr}, histopathology~\cite{qaiser2019fast}, brain functionality analysis~\cite{caputi2021promises}, fMRI data~\cite{rieck2020uncovering}, and genomic data~\cite{rabadan2019topological}. For a thorough review of TDA methods in biomedicine, see the excellent survey~\cite{skaf2022topological}. For a collection of TDA applications in various domains, see the TDA Applications Library~\cite{giunti22,DONUT}. 

Recently, there has been a surge of interest in topological deep learning within the machine learning community, showing great potential to augment existing deep learning methodologies~\cite{papamarkou2024position,zia2023topological}. The effective utilization of topological features has notably enhanced CNN models, particularly in tasks such as image segmentation~\cite{gupta2022learning,santhirasekaram2023topology,stucki2023topologically}. Moreover, there is growing recognition of the pivotal role of topological features in diagnostic tasks across various medical domains~\cite{somasundaram2021persistent,yadav2023histopathological}. To our knowledge, our work presents the first application of TDA methods for the detection of ovarian cancer.

\section{Methodology} \label{sec:methodology}

Our methodology consists of two primary steps. First, we extract the topological feature vectors from the images. Then, we apply these topological features directly to machine learning models, resulting in a fast and interpretable model called \textit{TopOC-1}. Next, we integrate these vectors with existing deep learning models for cancer diagnosis, resulting in \textit{TopOC-CNN}.

\vspace{-.1in}

\subsection{Topological Vectors for Images} \label{sec:PH_background}

Persistent homology (PH) serves as a robust mathematical tool within topological data analysis (TDA) for exploring the intricate shape and structure inherent in complex datasets. Its fundamental concept involves systematically examining the development of various hidden patterns within the data at different resolutions~\cite{chazal2021introduction}. PH demonstrates remarkable effectiveness in extracting features from diverse data formats such as point clouds and networks. However, our paper focuses exclusively on its application in image analysis, specifically emphasizing \textit{cubical persistence}, which is a specific version of PH.
While we aim to describe the construction of PH in accessible terms for non-specialists, those seeking deeper insights can refer to the comprehensive textbook~\cite{dey2022computational}. Essentially, PH can be summarized as a three-step procedure as follows.

\begin{enumerate}
    \item \textbf{Filtration}: Inducing a sequence of nested topological spaces from the data.
    \item \textbf{Persistence Diagrams}: Recording the evolution of topological changes within this sequence.
    \item \textbf{Vectorization}: Converting these persistence diagrams into vectors to be utilized in machine learning models.
\end{enumerate}

\begin{figure}[t] 
\centering
    	\includegraphics[width=\linewidth]{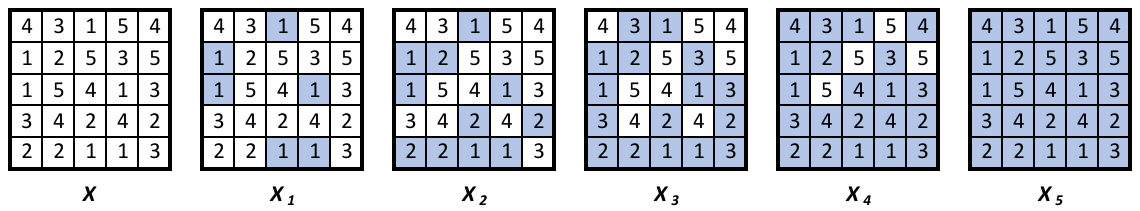}  
 {\caption{For the $5\times 5$ image $\X$ with the given pixel values, \textbf{the sublevel filtration} is the sequence of binary images $\X_1\subset \X_2\subset \X_3\subset \X_4\subset \X_5$.  \label{fig:filtration}}}
   \vspace{-.2in}
    \end{figure}

\paragraph{Step 1 - Constructing Filtrations.} Since PH essentially functions as a mechanism for monitoring the progression of topological characteristics within a sequence of simplicial complexes, constructing this sequence stands out as a crucial step. In image analysis, the common approach is to generate a nested sequence of binary images, also known as \textit{cubical complexes}. To achieve this from a given color (or grayscale) image $\X$ (with dimensions $r\times s$), one needs to select a specific color channel (e.g., red, blue, green, or grayscale). The color values $\gamma_{ij}$ of individual pixels $\Delta_{ij} \subset \X$ are then utilized. Specifically, for a sequence of color values ($0=t_1<t_2<\dots<t_N=255$), a nested sequence of binary images $\X_1\subset \X_2\subset\dots\subset \X_N$ is obtained, where $\X_n=\{\Delta_{ij} \subset \X \mid  \gamma_{ij}\leq t_n\}$ (See \Cref{fig:filtration}).

In particular, this involves starting with a blank $r\times s$ image and progressively activating (coloring black) pixels as their grayscale values reach the specified threshold $t_n$. This process, known as {\em sublevel filtration}, is conducted on $\X$ relative to a designated function (in this instance, grayscale). Alternatively, one can activate pixels in descending order, referred to as {\em superlevel filtration}. In other words, let $\Y_n=\{\Delta_{ij} \subset \X \mid  \gamma_{ij}\geq s_n\}$ for ($255=s_1>s_2>\dots > s_M=0$), and $\Y_1\subset \Y_2\subset \dots\subset \Y_M$ is called superlevel filtration.

\vspace{-.1in}

\paragraph{Step 2 - Persistence Diagrams.} PH traces the development of topological characteristics across the filtration sequence $\{\X_n\}$ and presents it through a \textit{persistence diagram} (PD). Specifically, if a topological feature $\sigma$ emerges in $\X_m$ and disappears in $\X_n$ with $1\leq m<n\leq N$, the thresholds $t_m$ and $t_n$ are denoted as the {\em birth time} $b_\sigma$ and {\em death time} $d_\sigma$ of $\sigma$, respectively ($b_\sigma=t_m$ and $d_\sigma=t_n$). Therefore, PD contains all such $2$-tuples $\PD_k(\X)=\{(b_\sigma,d_\sigma)\}$ where $k$ represents the dimension of the topological features. The interval $d_\sigma-b_\sigma$ is termed as the {\em lifespan} of $\sigma$. Formally, the $k^{th}$ persistence diagram can be defined as $\PD_k(\X)=\{(b_\sigma, d_\sigma) \mid \sigma\in H_k(\X_n) \mbox{ for } b_\sigma\leq t_n<d_\sigma\}$, where $H_k(\X_n)$ denotes the $k^{th}$ homology group of the cubical complex $\X_n$. Thus, $\PD_k(\X)$ contains $2$-tuples indicating the birth and death times of $k$-dimensional voids $\{\sigma\}$ (such as connected components, holes, and cavities) in the filtration sequence $\{\X_n\}$. For instance, for $\X$ in \Cref{fig:filtration}, $\PD_0(\X)=\{(1,\infty), (1,2), (1,2), (1,3)\}$ represents the connected components, while $\PD_1(\X)=\{(2,4),(3,5),(4,5)\}$ illustrates the holes in the corresponding binary images in \Cref{fig:filtration}.

% \begin{figure}[h!] 
% \centering
%     	\includegraphics[width=\linewidth]{figures/flowchart5.pdf}  
%  {\caption{Topological Vectors. We first generate persistence diagrams for any input image, utilizing grayscale values. Next, we derive our topological feature vectors, represented as Betti  functions. These vectors are then inputted into a simple ML classifier to produce classification results. \label{fig:flowchart1}}}
%     \end{figure}

\vspace{-.1in}

\paragraph{Step 3 - Vectorization.} Persistence Diagrams (PDs), which consist of collections of $2$-tuples, are not practical for use with ML tools. Instead, a common approach is to convert PD information into a vector or function, a process known as \textit{vectorization}~\cite{ali2023survey}. 

One commonly used function for this purpose is the \textit{Betti function}, which tracks the number of \textit{alive} topological features at each threshold. Specifically, the Betti function is a step function where $\beta_0(t_n)$ represents the count of connected components in the binary image $\X_n$, and $\beta_1(t_n)$ indicates the number of holes (loops) in $\X_n$. In the context of ML, Betti functions are typically represented as vectors $\overrightarrow{\beta_k}$ of size $N$ with entries $\beta_k(t_n)$ for $1\leq n\leq N$, defined as $\overrightarrow{\beta_k}(\X)=[ \beta_k(t_1)\  \beta_k(t_2)\ \dots \ \beta_k(t_N)]$.

For example, in \Cref{fig:filtration}, we can observe $\overrightarrow{\beta_0}(\X)=[4\  2\  1\  1\  1]$, indicating the count of connected components in the binary images $\{\X_i\}$, while $\overrightarrow{\beta_1}(\X)=[0\  1\  2\  2\  0]$ represents the counts of holes in $\{\X_i\}$. Notably, $\beta_0(1)=4$ signifies the count of components in $\X_1$, and $\beta_1(3)=2$ denotes the count of holes (loops) in $\X_3$.

There exist various other methods to convert PDs into a vector, such as persistence images~\cite{adams2017persistence}, persistence landscapes~\cite{bubenik2017persistence}, silhouettes~\cite{chazal2014stochastic}, and kernel methods~\cite{ali2023survey}. However, in this paper, we primarily utilize Betti functions due to their computational efficiency and ease of interpretation. %In~\Cref{fig:flowchart1}, we summarize how we utilize topological vectors directly with our straightforward ML model. Next, we describe how to integrate these  vectors with deep learning models.

\vspace{-.1in}

\subsection{TopOC Models} \label{sec:topoc}

%\BC{Saba, please create a new flowchart similar to \Cref{fig:flowchart1}.}

\begin{figure}[t] 
\centering
    	\includegraphics[width=\linewidth]{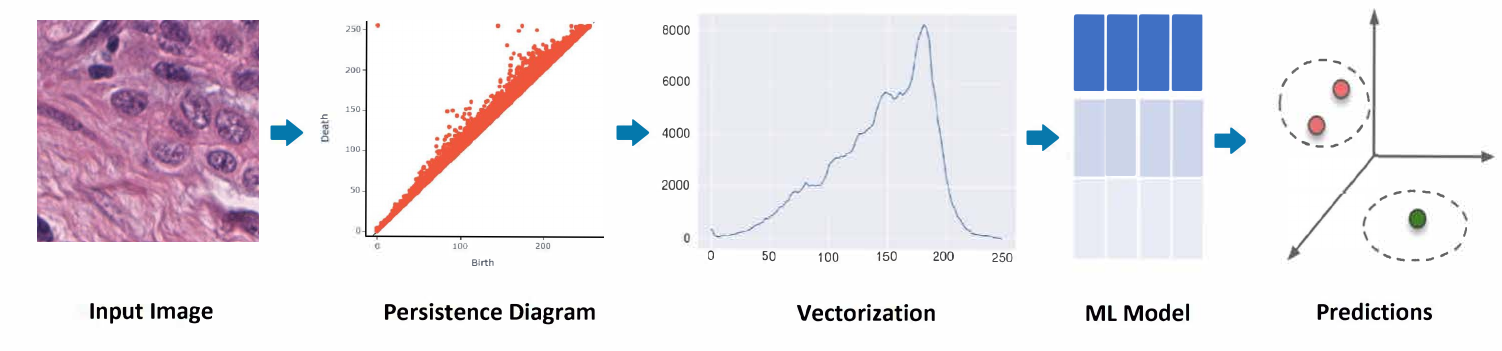}  
 {\caption{TopOC-1 Model. We first generate persistence diagrams for any input image, utilizing grayscale values. Next, we derive our topological feature vectors, represented as Betti functions. These vectors are then inputted into a simple ML classifier to produce classification results. \label{fig:flowchart1}}}
 \vspace{-.2in}
    \end{figure}

Once we have obtained the topological vectors, the next step involves applying ML tools to analyze the effectiveness of these features. Essentially, each medical image is represented as a 200-dimensional vector (300-dimensional for 3D images), signifying its embedding in a latent space. The classification task treats these embeddings as a point cloud and aims to identify distinct clusters corresponding to different classes. To evaluate the effectiveness of these topological vectors, we employed two different \textit{Topological Ovarian Cancer (TopOC)} models. 

\vspace{-.1in}

\paragraph{TopOC-1 model.} In our basic model, the TopOC-1 model (\Cref{fig:flowchart1}), we directly used ML classifier on topological embeddings of images to test the performance of topological vectors. The main idea is to use the procedure described in~\Cref{sec:PH_background}, for each image, we define a sublevel filtration with 50 thresholds to span [0, 255] color interval for each color. For an RGB image, we use Red, Green, Blue, and Grayscale (average of RGB color values) color values for each pixel, and we obtain four different sublevel filtrations one for each color. Next, we obtain the corresponding persistence diagrams, and by applying Betti vectorization, we obtain topological vectors for each image. Each color channel induces 50 dimensional $\beta_0(\X)$ and $\beta_1(\X)$. By concatenating these vectors, we obtain a 400-dimensional vector for each image. To eliminate correlations, we use dimension reduction by using feature selection methods. In our first model, TopOC-1, we simply apply standard ML classifiers on these vectors. Next, in our more sophisticated method, we integrate these vectors with deep learning methods.

%we used two types of ML classifiers: The first is eXtreme Gradient Boosting (XGBoost), a tree-based method renowned for its performance. The second is Multilayer Perceptron (MLP), a feedforward neural network that has gained popularity due to its adaptability and robustness. Both of these ML models offer a broad range of capabilities for classification tasks, especially in high-dimensional scenarios. MLP tends to excel when dealing with large training datasets and imbalanced datasets, while XGBoost consistently demonstrates good performance across various settings. In \Cref{sec:experiments}, we provide detailed information regarding our ML procedures and the process of hyperparameter tuning.

%\BC{Brighton, please add a flowchart for deep learning model similar to earlier ones.}

 \begin{figure}[t]
\centering
        \includegraphics[width=.8\textwidth]{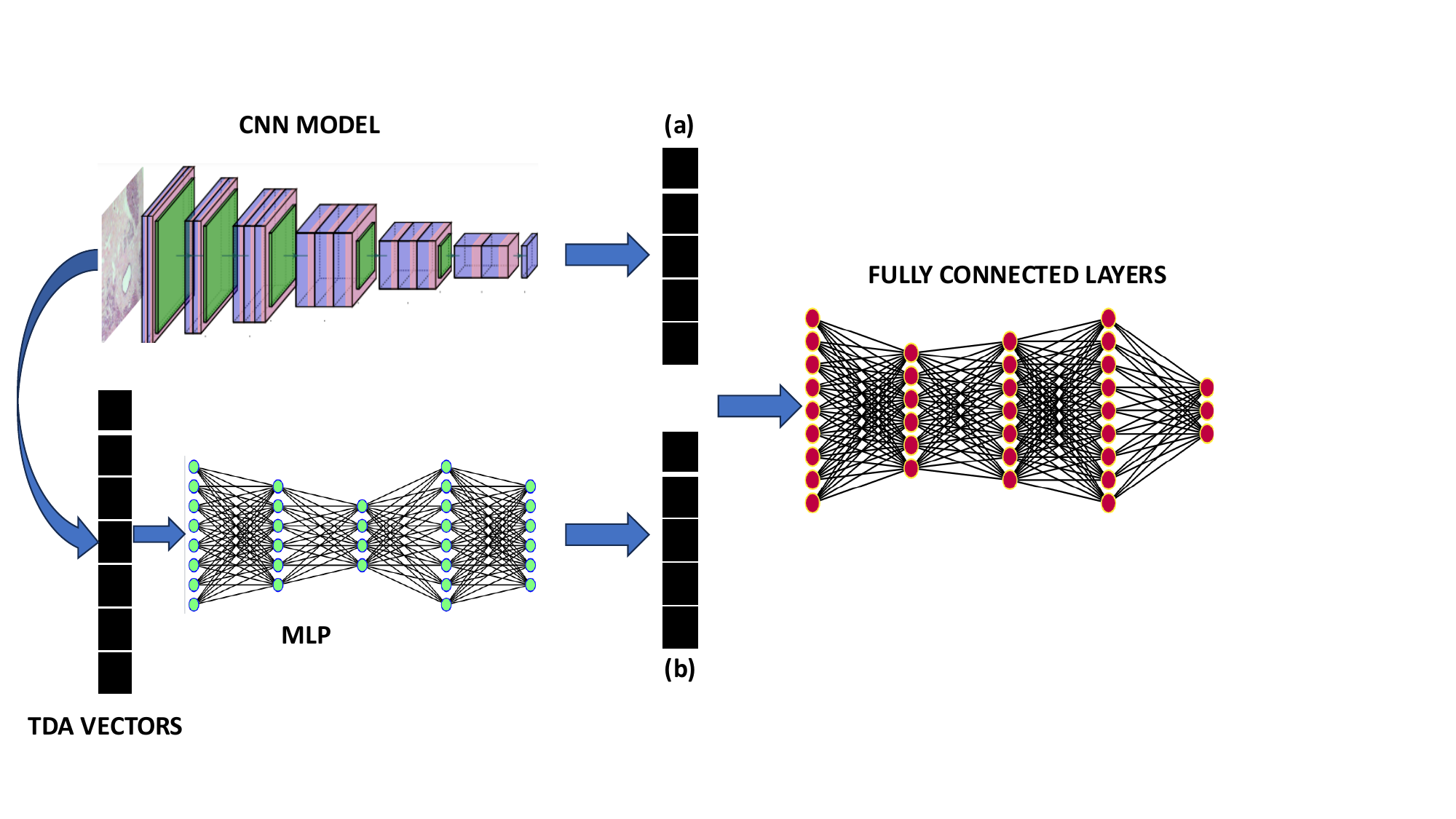}
    {\caption{\footnotesize {\bf TopOC-CNN Model.} In this model, we integrate topological features of images with convolutional vectors from a CNN backbone, followed by a fully connected layer. For the prediction head, we concatenate (a) a 128-dimensional CNN vector and (b) a topological vector, with dimension options of 256, 128, 64, or 0 (Vanilla-CNN). \label{fig:flowchart-CNN}}}
    %\BC{Brighton, please update the figure} \BN{Done, kindly let me know if there are adjacements I need to make}
    \vspace{-.2in}
\end{figure}

\vspace{-.1in}

\paragraph{TopOC-CNN model.} %\BC{Brighton, please revise} 
%\BN{Done, kindly read through and let me know if there is anything you would like me to readjust}
In our second model, TopOC-CNN  (\Cref{fig:flowchart-CNN}), we integrated topological features with pre-trained CNN models. This integration is based on the concept that topological features capture global patterns within the image, while convolutional vectors focus on more localized aspects. Combining these two distinct sets of features is expected to enhance the model's overall performance.
The TopOC-CNN model is meticulously crafted by selecting a base model from a range of state-of-the-art CNN models, including EfficientNetB0, DenseNet121, and VGG16. These models were chosen for their robust feature extraction capabilities, leveraging pre-trained weights from extensive datasets like ImageNet. The layers of these pre-trained CNN models remain frozen, and their capabilities are augmented by adding extra layers to enhance their ability to capture intricate patterns and features.
The CNN component of our model starts with an input layer that accepts images with a shape of $(224, 224, 3)$. This input is processed by the base model, whose output is then fed into an additional convolutional layer containing 256 filters, each with a $3$x$3$ kernel size and ReLU activation function. Subsequently, a max pooling layer with a $2$x$2$ pool size is applied to downsample the feature maps. These feature maps are then flattened into a single vector, which is further refined by a dense layer with $128$ neurons and a ReLU activation function.
Simultaneously, a Multilayer Perceptron (MLP) is defined with three dense layers, each utilizing the Rectified Linear Unit (ReLU) activation function. The MLP layers consist of $256$ neurons each, with the number of neurons in the last layer varying for different settings: $256$, $128$, $64$, and $0$ neurons. The outputs of both the augmented CNN layers ($128$ neurons) and the MLP ($256$, $128$, $64$, or $0$ neurons) are concatenated and passed through additional dense layers ($256$ neurons, $128$ neurons, $128$ neurons), all with ReLU activation functions, ultimately culminating in the output layer.

\vspace{-.1in}

\section{Experiments} \label{sec:experiments}

\subsection{Experimental Setup}

\noindent {\em Datasets.}\quad  To validate the performance of our models, we used two publicly available histopathology datasets. %UBC-OCEAN is a publicly available dataset, accessible at the Kaggle link~\cite{UBC-OCEAN}.
%\footnote{\url{https://www.kaggle.com/competitions/UBC-OCEAN}}. 
The first dataset, UBC-OCEAN, was provided in the UBC ovarian cancer subtype classification and outlier detection (UBC-OCEAN) challenge hosted by the University of British Columbia~\cite{asadi2024machine,farahani2022deep}. The dataset is accessible at the Kaggle link~\cite{UBC-OCEAN} and provides 34K tiles of 224$\times$224 size obtained from 975 whole slide images (WSI) and tissue microarrays (TMA). The task is to classify five subtypes of ovarian cancer, namely clear cell ovarian carcinoma (CC), endometrioid (EC), high-grade serous carcinoma (HGSC), low-grade serous carcinoma (LGSC), and mucinous carcinoma (MC). We used the pre-defined training and test split (91:9) provided by the challenge, i.e., $31,203$ training and $3,082$  test images.
 Our second dataset, BREAKHIS, is the Breast Cancer Histopathology dataset~\cite{breakhis2024}, consisting of 7909 samples with two classes, Benign and Malignant. These samples are provided at 700$\times$460 size with different magnification factors such as 40x, 100x, 200x, and 400x~\cite{spanhol2016breast,spanhol2017deep}. For this dataset, we used a commonly employed 70:30 training-test split~\cite{benhammou2020breakhis}.

% OTU-2D is part of publicly available MMOTU dataset containing 1469 2D Ovarian Tumor Ultrasound images collected from 294 patients with two classes(benign and Malignant)~\cite{zhao2022multi}. 

%\BC{Saba, please work on a draft for experimental setup similar to histopathology paper.}

\smallskip

\noindent {\em Topological Vectors.}\quad As described in~\Cref{sec:PH_background}, for each image, by employing a filtration for each color channel (R, G, B, and G), we obtain four filtration. We employed 50-thresholds, and each color channel, produces 50-dimensional $\overrightarrow{\beta_0}$ and $\overrightarrow{\beta_1}$ vectors (See~\Cref{fig:betti-curves1}). Next, by concatenating them, we got a 400-dimensional topological vector for each image. 

In \Cref{sec:exp-appendix}, we give further experimental details, including \textit{Model Hyperparameters, Extra Performance Metrics,} and \textit{Computational Complexity and Runtime}.

%\paragraph{Feature Selection.} To increase the performance of our TopOC-1 model, we employed a feature selection method. However, some of the color channels may not produce relevant information for downstream tasks, or some features may be correlated. We employed XGBoost as ML Classifier, and \textit{feature\_importance}, a built-in function in XGBoost library, ranks features with respect to their importance. Out of 400-features, we tried 50, 100, 200, and 400 best features for our models~\Cref{tab:feature-selection}. We observed the best performance with 200 features, and utilized this choice in our TopOC-1 model.

\vspace{-.1in}

\subsection{Results}

%\BC{After adding accuracy tables, we'll write this part.}

\begin{wraptable}{r}{6cm}
\vspace{-.25in}
\centering
    \setlength\tabcolsep{4pt}
\caption{\footnotesize {\bf TopOC-1.} Performances of our TopOC-1 model with different numbers of topological features.\label{tab:feature-selection}}
\resizebox{\linewidth}{!}{
\begin{tabular}{lccc|cccc}
\toprule
& \multicolumn{3}{c}{\textbf{UBC-OCEAN}} & \multicolumn{3}{c}{\textbf{BREAKHIS-40x}} \\
\cmidrule(lr){2-4} \cmidrule(lr){5-7}  
\textbf{Models} & \textbf{B. Acc.} & \textbf{Acc.} & \textbf{AUC} & \textbf{Acc.} & \textbf{Sen.} & \textbf{Spec.}  \\
\midrule
50 Features &61.14 & 67.32&89.07 &87.81 &94.51 &  74.24\\
100 Features &63.45 &69.92 &90.94 &89.32 & 96.51&74.75  \\
200 Features &\textcolor{blue}{\bf 66.13} &\textcolor{blue}{\bf 72.45} & 91.74&\textcolor{blue}{\bf 90.82} & \textcolor{blue}{\bf 97.76}& \textcolor{blue}{\bf 76.77} \\
400 Features &65.31 &71.51 & \textcolor{blue}{\bf 91.75}&89.98 & 97.76& 74.24 \\
\bottomrule
\end{tabular}}
\vspace{-.25in}
\end{wraptable}

UBC-OCEAN is a recent challenge that concluded a few months ago, offering a \$50K prize. The performance metric was balanced accuracy, and the winning entry achieved 66\% balanced accuracy for this five-class classification task~\cite{UBC-OCEAN}. Both of our models surpassed this benchmark. TopOC-1, using 200 features, achieved 66.13\% balanced accuracy as shown in~\Cref{tab:feature-selection}. Our TopOC-CNN model, employing a DenseNet121 backbone and 128-dimensional topological vectors, reached 67.15\% balanced accuracy. Notably, we utilized only pre-trained models without any data augmentation, whereas many competitors relied on customized CNN models with extensive data augmentation and other optimizations.

\begin{table}[t]
\vspace{-.2in}
\centering
    \setlength\tabcolsep{4pt}
\caption{\footnotesize Balanced Accuracy, Accuracy, and AUC performances for TopOC-CNN models with different backbones and varying topological MLP final output layer settings (64, 128, 256) on the multiclass classification of UBC-OCEAN dataset.\label{tab:topo-CNN}}
\resizebox{.8\textwidth}{!}{
\begin{tabular}{lccc|ccc|ccc}
\toprule
& \multicolumn{3}{c}{\textbf{DenseNet121}} & \multicolumn{3}{c}{\textbf{EfficientNetB0}} & \multicolumn{3}{c}{\textbf{VGG16}} \\
\cmidrule(lr){2-4} \cmidrule(lr){5-7}  \cmidrule(lr){8-10}
\textbf{Models} & \textbf{B. Acc.} & \textbf{Acc.}  & \textbf{AUC} & \textbf{B. Acc.} & \textbf{Acc.}  & \textbf{AUC} & \textbf{B. Acc.} & \textbf{Acc.}  & \textbf{AUC} \\
\midrule
Vanilla-CNN& 65.10 & 69.08  & 89.42 & 56.08 & 63.82 & 87.04 & 56.45 & 61.55  & 84.02 \\
\midrule
CNN+64Top & 64.62 & 68.59  & 89.37 & 59.85 & 64.34  & 86.71 & 56.50 & 61.26  & 83.53\\
CNN+128Top & \textcolor{blue}{\bf 67.15} & 69.34  & \textcolor{blue}{\bf 92.08} & 59.96 & 64.05  & 86.48 & 55.00 & 62.20  & \textcolor{blue}{\bf 84.33}\\
CNN+256Top & 66.36 & \textcolor{blue}{\bf 69.99}  & 89.62 & \textcolor{blue}{\bf 62.67} & \textcolor{blue}{\bf 67.98}  & \textcolor{blue}{\bf 89.14} & \textcolor{blue}{\bf 57.63} & \textcolor{blue}{\bf 63.24}  & 84.24 \\
\midrule
    Improvement & \textcolor{blue}{2.05}	&\textcolor{blue}{0.91} &	\textcolor{blue}{2.66}	& \textcolor{blue}{6.59}	& \textcolor{blue}{4.16} 	& \textcolor{blue}{2.10} &	\textcolor{blue}{1.18}	& \textcolor{blue}{1.69}  & \textcolor{blue}{0.31}\\
\bottomrule
\end{tabular}}
\vspace{-.2in}
\end{table}

\begin{table}[t]
\centering
    \setlength\tabcolsep{4pt}
\caption{\footnotesize Performance comparison of Vanilla-CNN and TopOC-CNN models with different backbones for BREAKHIS dataset.\label{tab:topo-CNN-breakhis}}
\resizebox{.8\textwidth}{!}{
\begin{tabular}{llccc|ccc|ccc}
\toprule
& & \multicolumn{3}{c}{\textbf{DenseNet121}} & \multicolumn{3}{c}{\textbf{EfficientNetB0}} & \multicolumn{3}{c}{\textbf{VGG16}} \\
 \cmidrule(lr){3-5} \cmidrule(lr){6-8}  \cmidrule(lr){9-11}
\textbf{Mag.} & \textbf{Models} & \textbf{Acc} & \textbf{Sens.}  & \textbf{Spec} & \textbf{Acc} & \textbf{Sens}  & \textbf{Spec}& \textbf{Acc} & \textbf{Sens}  & \textbf{Spec}  \\
\midrule
\multirow{3}{*}{\bf 40x} &Vanilla-CNN&92.99  & 91.72 & 88.30 &  84.81  &84.46  &83.51 & 88.81 & 87.38  & 83.51 \\
& CNN+256Top &   94.82  &   94.06 &  92.02  & 93.16 & 91.12 &  85.64& 90.48  & 89.31  &86.17 \\
%\midrule
& Improvement & \textcolor{blue}{1.83} &	\textcolor{blue}{2.34} &	\textcolor{blue}{3.72} &	\textcolor{blue}{8.35}&	\textcolor{blue}{6.66}&	\textcolor{blue}{2.13} &	\textcolor{blue}{1.67} &	\textcolor{blue}{1.93} &	\textcolor{blue}{2.66}
\\
\midrule
\multirow{3}{*}{\bf 100x} &Vanilla-CNN & 90.40   &88.18  & 82.38 &  89.12  & 85.68 & 76.68   & 87.84  &  84.18 & 74.61  \\
& CNN+256Top & 92.48   &90.98  & 87.05 & 92.16 & 91.03   & 88.08  &  92.32  & 90.57  & 86.01\\
%\midrule
& Improvement &\textcolor{blue}{2.08}&\textcolor{blue}{2.80}&\textcolor{blue}{4.67}&\textcolor{blue}{3.04}&\textcolor{blue}{5.35}&\textcolor{blue}{11.4}&\textcolor{blue}{4.48}&\textcolor{blue}{6.39}&\textcolor{blue}{11.4} \\
\midrule
\multirow{3}{*}{\bf 200x} &Vanilla-CNN & 91.72   &90.17  & 86.10 & 87.58   & 84.08 &  74.86   & 87.58   &  85.40 &  79.68  \\
& CNN+256Top & 93.71   & 92.79  & 90.37  & 93.71  &    91.76 & 86.63  &  90.07   & 88.23  &  83.42 \\
%\midrule
& Improvement &\textcolor{blue}{1.99}&\textcolor{blue}{2.62}&\textcolor{blue}{4.27}&\textcolor{blue}{6.13}&\textcolor{blue}{7.68}&\textcolor{blue}{11.77}&\textcolor{blue}{2.49}&\textcolor{blue}{2.83}&\textcolor{blue}{3.74} \\
\midrule
\multirow{3}{*}{\bf 400x} &Vanilla-CNN & 91.03   & 88.16 & 80.11 & 85.35   &81.74  &  71.59   & 85.53  & 81.88 &  71.59  \\
&CNN+256Top & 93.22  &  92.17 &  89.20& 90.29  &  87.18  & 78.41  & 87.73  & 84.84 & 76.70 \\
%\midrule
& Improvement &\textcolor{blue}{2.19}&\textcolor{blue}{4.01}&\textcolor{blue}{9.09}&\textcolor{blue}{4.94}&\textcolor{blue}{5.44}&\textcolor{blue}{6.82}&\textcolor{blue}{2.20}&\textcolor{blue}{2.96}&\textcolor{blue}{5.11} \\
\bottomrule
\end{tabular}}
\vspace{-.25in}
\end{table}

\begin{table}[b]
\vspace{-.3in}
\caption{\footnotesize Performance comparison of TopOC model with deep learning models on binary classification of BREAKHIS dataset. \label{table:Breakhis1}}
\centering
    \setlength\tabcolsep{6pt}
\resizebox{\textwidth}{!}{%
\begin{tabular}{lccc|ccc|ccc|ccc}
\toprule
& \multicolumn{3}{c}{\textbf{40x}} & \multicolumn{3}{c}{\textbf{100x}} & \multicolumn{3}{c}{\textbf{200x}} & \multicolumn{3}{c}{\textbf{400x}} \\
\cmidrule(lr){2-4} \cmidrule(lr){5-7}  \cmidrule(lr){8-10} \cmidrule(lr){11-13} 
\textbf{Method} & \textbf{Acc.} & \textbf{Sens.} & \textbf{Spec.} & \textbf{Acc.} & \textbf{Sens.} & \textbf{Spec.} & \textbf{Acc.} & \textbf{Sens.} & \textbf{Spec.} & \textbf{Acc.} & \textbf{Sens.} & \textbf{Spec.} \\ 
\midrule
%NDCNN [21] & 82 & – & – & 86.2 & – & – & 84.6 & – & – & 84 & – & – \\
%NPMIL [35] & 87.8 & – & – & 85.6 & – & – & 80.8 & – & – & 82.9 & – & – \\
AlexNet~\cite{spanhol2016breast} & 81.52 & 75.64 & 87.40 & 81.28 & 78.16 & 84.40 & 83.54 & 79.16 & 87.91 & 81.10 & 76.28 & 85.90 \\
BkNet~\cite{wang2021automatic} & 85.61 & 84.42 & 86.80 & 86.23 & 87.19 & 85.26 & 85.37 & 80.01 & 90.74 & 84.43 & 80.00 & 88.87 \\
CapsNet~\cite{sabour2017dynamic} & 86.95 & 86.29 & 87.61 & 89.13 & 88.30 & \textcolor{blue}{\underline{89.96}} & 88.75 & 86.22 & \textcolor{blue}{\underline{91.28}} & 88.04 & 87.56 & 88.51 \\
%E-CapsNet & 91.67 & 90.09 & 93.25 & 93.87 & 94.36 & 93.39 & 93.34 & 93.64 & 93.04 & 92.85 & 92.69 & 93.01 \\
BkCapsNet~\cite{wang2021automatic} & \textcolor{blue}{\underline{92.71}} &92.15&  \textcolor{blue}{\bf 93.27} &\textcolor{blue}{\bf 94.52} & \textcolor{blue}{\underline{95.16}} & \textcolor{blue}{\bf 93.87} &  \textcolor{blue}{\bf 94.03} &\textcolor{blue}{\underline{94.31}} &  \textcolor{blue}{\bf 93.75} &  \textcolor{blue}{\bf 93.54}& \textcolor{blue}{\underline{94.06}} &  \textcolor{blue}{\bf 93.03} \\ 
\midrule
TopOC-1 & 90.82 & \textcolor{blue}{\bf 97.76} & 76.77 &91.68 & \textcolor{blue}{\bf 96.00} & 82.50 &91.05 &  \textcolor{blue}{\bf 96.53}& 80.00 &88.64 &\textcolor{blue}{\bf 96.02} &75.25 \\
TopOC-CNN &   \textcolor{blue}{\bf 94.82}  &   \textcolor{blue}{\underline{94.06}} &  \textcolor{blue}{\underline{92.02}} &\textcolor{blue}{\underline{92.48}} &91.03 &88.08 &\textcolor{blue}{\underline{93.71}} &92.79& 90.37&\textcolor{blue}{\underline{93.22}}&92.17&\textcolor{blue}{\underline{89.20}}\\
\bottomrule
\end{tabular}}
\end{table}

% \begin{table}[ht]
% \centering
% \caption{Performance metrics for different TopOC-CNN models with varying MLP final output layer settings (256, 128, 64, or 0 neurons)}
% \begin{tabular}{lcccc}
% \toprule
% \textbf{Model} & \textbf{Bal. Acc.} & \textbf{Precision} & \textbf{Recall} & \textbf{AUC} \\
% \midrule
% DenseNet121 (0) & 65.10 & 69.08 & 69.08 & 89.42 \\
% DenseNet121 (64) & 64.62 & 68.59 & 68.59 & 89.37 \\
% DenseNet121 (128) & 67.15 & 69.34 & 69.34 & 92.08 \\
% DenseNet121 (256) & 66.36 & 69.99 & 69.99 & 89.62 \\
% \midrule
% EfficientNetB0 (0) & 56.08 & 63.82 & 63.82 & 87.04 \\
% EfficientNetB0 (64) & 59.85 & 64.34 & 64.34 & 86.71 \\
% EfficientNetB0 (128) & 59.96 & 64.05 & 64.05 & 86.48 \\
% EfficientNetB0 (256) & 62.67 & 67.98 & 67.98 & 89.14 \\
% \midrule
% ResNet101 (0) & 65.82 & 69.57 & 69.56 & 89.31 \\
% ResNet101 (64) & 66.50 & 69.95 & 69.95 & 89.42 \\
% ResNet101 (128) & 63.61 & 68.59 & 68.59 & 88.61 \\
% ResNet101 (256) & 66.89 & 69.99 & 69.99 & 89.87\\ 
% \midrule
% ResNet50 (0) & 66.45 & 68.92 & 68.92 & 88.85 \\
% ResNet50 (64) & 64.99 & 69.66 & 69.66 & 89.44 \\
% ResNet50 (128) & 65.37 & 67.68 & 67.68 & 87.90 \\
% ResNet50 (256) & 65.79 & 68.88 & 68.88 & 88.76 \\
% \midrule
% VGG16 (0) & 56.45 & 61.55 & 61.55 & 84.02 \\
% VGG16 (64) & 56.50 & 61.26 & 61.26 & 83.53 \\
% VGG16 (128) & 55.00 & 62.20 & 62.20 & 84.33 \\
% VGG16 (256) & 57.63 & 63.24 & 63.24 & 84.24 \\
% \bottomrule
% \end{tabular}
% \label{tab:performance-metrics}
% \end{table}

In~\Cref{tab:topo-CNN}, we present the TopOC-CNN results for three backbones (DenseNet121, EfficientNetB0, VGG16) and four variations of each. Specifically, in~\Cref{fig:flowchart-CNN}, the CNN vector (a) is fixed at 128-dimension across all models, while the topological vector (b) (MLP output) varies among 0 (Vanilla-CNN), 64, 128, and 256-dimension. Our findings indicate that incorporating topological vectors consistently enhances CNN performance, up to 6.59\% improvements in balanced accuracy.

For the BREAKHIS dataset, both of our models also deliver highly competitive results in breast cancer diagnosis~(\Cref{table:Breakhis1}). Similarly, our topological vectors enhance CNN model performance by up to 8.35\% in accuracy~(\Cref{tab:topo-CNN-breakhis}). Additional performance metrics for our models are provided in~\Cref{table:BREAKHIS2}.

We provide our model's \textit{Visualization and Interpretability} discussion in \Cref{sec:interpret}.

%\BN{Why donot we use different colors for the best and second best results, it is quite confusing between bold blue and the normal blue underlined in table \ref{table:Breakhis1}}

\vspace{-.1in}
\section{Conclusion}
\vspace{-.05in}

In conclusion, our study underscores the substantial promise of integrating topological machine learning methods with existing deep learning techniques for histopathological image analysis in ovarian and breast cancers. The developed models, TopOC-1 and TopOC-CNN, exhibit enhanced accuracy and efficiency in cancer detection by incorporating topological features into the deep learning framework. Moving along, we aim to focus on expanding model validation across diverse datasets and cancer types, refining topological feature extraction techniques, and incorporating these methods into real-time clinical diagnostic tools. Addressing the challenge of obtaining large labeled datasets through collaboration and advanced learning techniques will further optimize these models, ultimately advancing more accurate and accessible cancer diagnostics.

\begin{credits}
\subsubsection{\ackname} This work was partially supported by the National Science Foundation under grants  DMS-2202584, 2229417, and DMS-2220613 and by Simons Foundation under grant \# 579977.
The authors acknowledge the \href{http://www.tacc.utexas.edu}{Texas Advanced Computing Center} (TACC) at UT Austin for computational resources which contributed to the research results reported within this paper. 
\end{credits}

\bibliographystyle{splncs04}
\bibliography{references}

\clearpage

\setcounter{page}{1}

\appendix

\centerline{\large \bf Appendix}

\section{Experimental Details} \label{sec:exp-appendix}

% \begin{table}[h!]
% \vspace{-.1in}
% \caption{Subtypes and Patch Counts for UBC-OCEAN (Ovarian) and BREAKHIS (Breast) Histopathology Datasets. \label{tab:datasets}}
% \centering
%     \setlength\tabcolsep{6pt}
% \resizebox{.7\textwidth}{!}{
% \begin{tabular}{ccccc|cc}
% \toprule
% \multicolumn{5}{c}{\textbf{UBC-OCEAN}} & \multicolumn{2}{c}{\textbf{BREAKHIS}}  \\
% \cmidrule(lr){1-5} \cmidrule(lr){6-7}  
% \textbf{CC} & \textbf{EC} & \textbf{HGSC} & \textbf{LGSC} & \textbf{MC} & \textbf{Benign} & \textbf{Malignant} \\ 
% \midrule
% 6130 & 8154& 13207 & 3195 & 3599 & 2480& 5429 \\
% \bottomrule
% \end{tabular}}
% \end{table}

\smallskip

% \BN{Dear Saba, kindly read through the following (Hyperparameters) and let me know if there is another I need to readjust.}
\noindent {\em Hyperparameters.}\quad
For the TopOC-1 model, we trained an XGBoost model on the datasets, employing specific hyperparameters and feature selection techniques. For XGBoost, parametric tuning involved setting the learning rate to $0.05$, the maximum depth to $5$, the number of estimators to $300$, the subsample to $0.9$, and the colsample bytree to $0.9$. To increase the performance of our TopOC-1 model by removing correlated features, we employed a feature selection method. %However, some of the color channels may not produce relevant information for downstream tasks, or some features may be correlated. 
We employed XGBoost as ML Classifier, and \textit{feature\_importance}, a built-in function in the XGBoost library, ranks features with respect to their importance. Out of 400 features, we tried 50, 100, 200, and 400 best features for our models~\Cref{tab:feature-selection}. We observed the best performance with 200 features and utilized this choice in our TopOC-1 model.

%For feature selection, we utilized the feature importance scores generated by the XGBoost model. These scores were used to rank features in descending order of importance, and the top $200$ features were selected for model training. Features with importance scores below the threshold of the top $200$ were considered unimportant and removed.

For the TopOC-CNN model, the model is configured with a set of well-defined hyperparameters to optimize its performance. We use the Adam optimizer for its efficiency and adaptability. The loss function selected is categorical cross-entropy, suitable for multi-class classification tasks or binary cross-entropy for binary classification. The batch size is set to $64$, balancing computational efficiency and convergence speed. The training process is conducted over $100$ epochs, allowing sufficient iterations for the model to learn from the data. %Key performance metrics, including accuracy, precision, recall, and Area Under the Curve (AUC), are incorporated to comprehensively evaluate the model's predictive capabilities and ensure robust performance. 
%\BC{Brighton, create a link, and add it here. https://anonymous.4open.science/r/Ovarian-and-Breast-Cancer-Diagnosis-8F2C}
We shared our code at the following link: {\footnotesize \url{https://anonymous.4open.science/r/Ovarian-and-Breast-Cancer-Diagnosis-8F2C}}.

\smallskip

\noindent {\em Performance Metrics.} \quad We used a variety of metrics to evaluate the efficiency of our models, including accuracy, balanced accuracy, precision, sensitivity (recall), specificity, and the F-1 score~\cite{hicks2022evaluation}. For multiclass evaluation in the UBC-OCEAN challenge, the performance metric utilized was \textit{balanced accuracy}, which is the macro average of the recall obtained in each class. Therefore, for a balanced dataset, balanced accuracy tends to be the same as accuracy. However, by giving equal importance to each class, balanced accuracy is considered more suitable for unbalanced datasets~\cite{grandini2020metrics}.

\begin{table}[b]
\vspace{-.2in}
\caption{Further performance metrics for TopOC-1 model on BREAKHIS dataset for binary classification. \label{table:BREAKHIS2}}
\centering
    \setlength\tabcolsep{6pt}
\resizebox{\textwidth}{!}{%
\begin{tabular}{lccccccccc}
\toprule
\textbf{Mag.} & \textbf{Accuracy} & \textbf{Sensitivity} & \textbf{Specificity} & \textbf{AUC} & \textbf{Precision} & \textbf{Recall} & \textbf{F1-Score} \\ 
\midrule
40x &  90.82 & 97.76 & 76.77&96.18 & 89.50& 97.76& 93.44& \\
100x &91.68 & 96.00 & 82.50 &95.95 &92.10 &96.00 &94.01 \\
200x &91.05 &  96.53 & 80.00  &94.95 &90.69 &96.53 &93.52  \\
400x &88.64 & 96.02 &75.25&94.77 &87.56 &96.02 &91.59 & \\
\bottomrule
\end{tabular}}
\vspace{-.2in}
\end{table}

\smallskip

%\BN{professor, I added the following paragraph, kindly read through}
\noindent {\em Computational Complexity and Runtime.}\quad
PH calculation can be computationally demanding, especially in high-dimensional data scenarios~\cite{otter2017roadmap}. However, when dealing with image data, it exhibits notable efficiency. Specifically, for 2D images, the time complexity of PH is approximately $\mathcal{O}(|\mathcal{P}|^r)$, where $r$ is approximately 2.37, and $|\mathcal{P}|$ represents the total number of pixels~\cite{milosavljevic2011zigzag}. In simpler terms, the computational effort required for PH increases almost quadratically with the image size. It's important to note that the remaining processes such as vectorization and ML steps are significantly minor in terms of computational load when compared to the PH step.

In our experiments, it took an average of 5 hours and 16 minutes to generate topological vectors for all color channels for the UBC-OCEAN dataset, and our TopOC-CNN model on average trained for $11$ hours and $55$ minutes. For the BreaKHis dataset, it took an average of $50$ minutes to generate the vectors, and an average of $1$ hour and $2$ minutes to train our TopOC-CNN model.  
All experiments were conducted using high-performance computing clusters with the following SLURM script configuration: partition=gpu-a100-small, nodes=$1$, and ntasks=$1$. Each node was equipped with 3x NVIDIA A100 GPUs (40GB HBM2), 2x AMD EPYC 7763 64-core Processors, 128 total cores per node, and 256 GB of RAM.

\vspace{-.15in}

\section{Visualization and Interpretability} \label{sec:interpret}

\begin{figure}[t]
    \centering
         \begin{subfigure}[b]{0.32\textwidth}
         \centering
         \includegraphics[width=\textwidth]{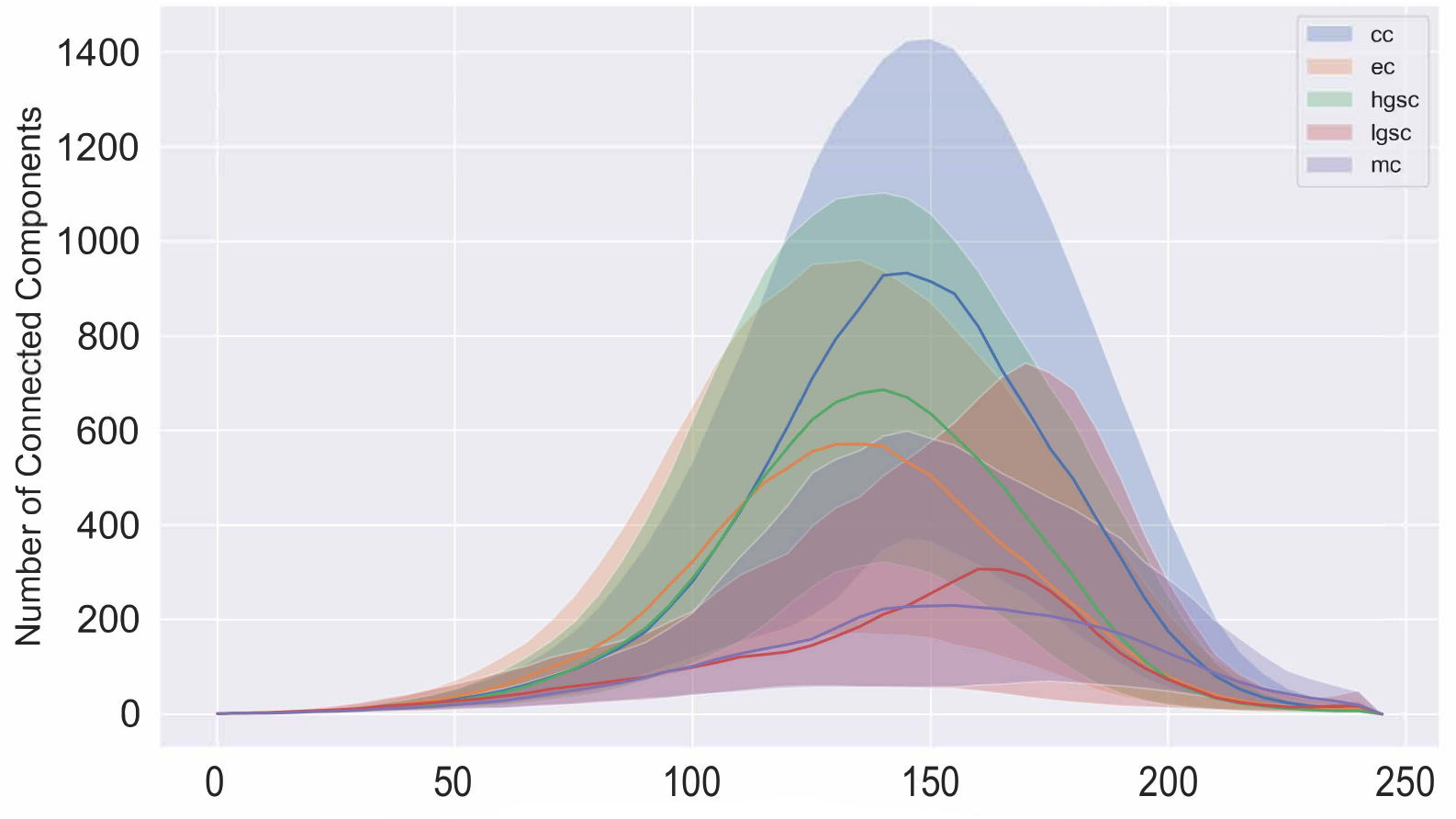}
         \caption{Betti-0 (Gray)}
         \label{fig:Betti-0 (gray)}
     \end{subfigure}
    % \begin{subfigure}[b]{0.49\textwidth}
    %      \centering
    %      \includegraphics[width=\textwidth]{figures/cc-ec_grayscale_betti0_50.pdf}
    %      \caption{Betti-0 curves for grayscale\\(cc vs. ec)}
    %      \label{fig:Betti-0-gray (cc-ec)}
    %  \end{subfigure}
    %  \hfill
    %  \begin{subfigure}[b]{0.49\textwidth}
    %      \centering
    %      \includegraphics[width=\textwidth]{figures/hgsc-lgsc_grayscale_betti0_50.pdf}
    %      \caption{Betti-0 curves for grayscale\\(hgsc vs. lgsc)}
    %      \label{fig:Betti-0-gray (hgsc-lgsc)}
    % \end{subfigure}
     \begin{subfigure}[b]{0.33\textwidth}
         \centering
         \includegraphics[width=\textwidth]{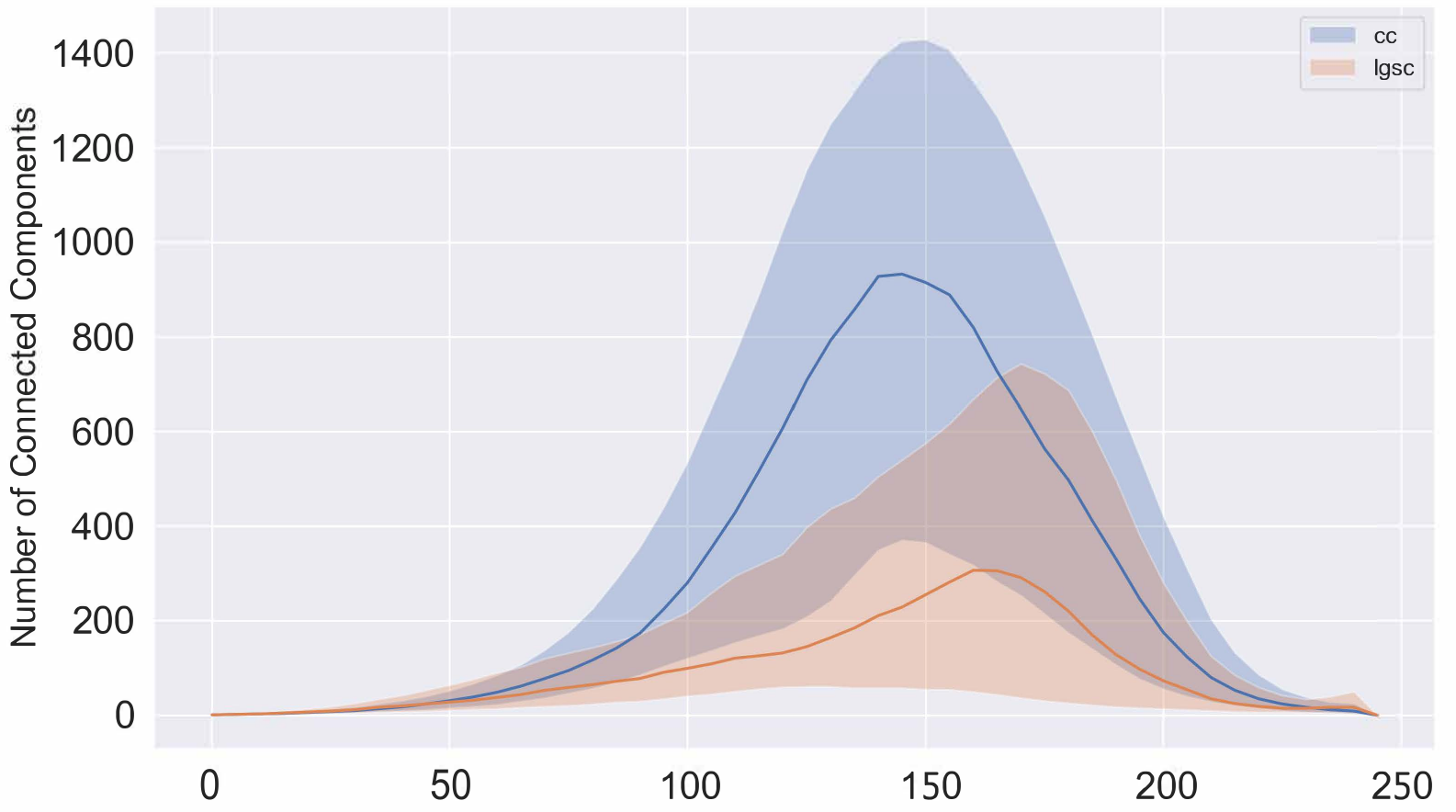}
         \caption{CC vs. LGSC}
         \label{fig:Betti-0-gray (cc-lgsc)}
    \end{subfigure}
     \begin{subfigure}[b]{0.33\textwidth}
         \centering
         \includegraphics[width=\textwidth]{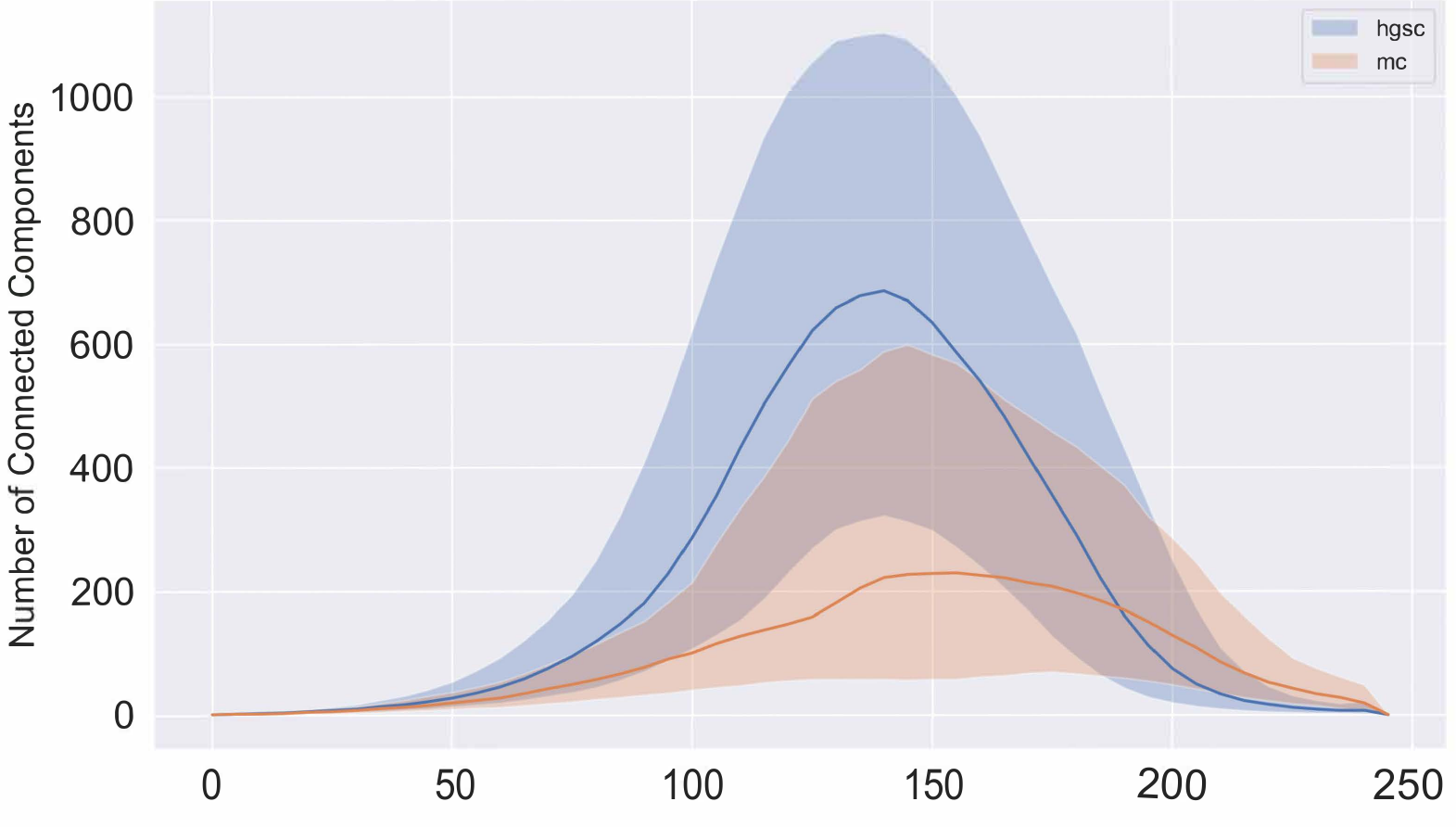}
         \caption{HGSC vs. MC}
         \label{fig:Betti-0-gray (hgsc vs. mc)}
    \end{subfigure}
         \caption{\footnotesize In the left figure, we present the median curves and 40\% confidence bands of Betti-0 (grayscale) curves for five classes from the UBC-OCEAN dataset. In the next two figures, we display the same curves for two specific classes.          \label{fig:betti-curves1}}
         \vspace{-.2in}
\end{figure}

\begin{figure}[h!]
\vspace{-.23in}
     \centering
     \begin{subfigure}[b]{0.32\textwidth}
         \centering
         \includegraphics[width=\textwidth]{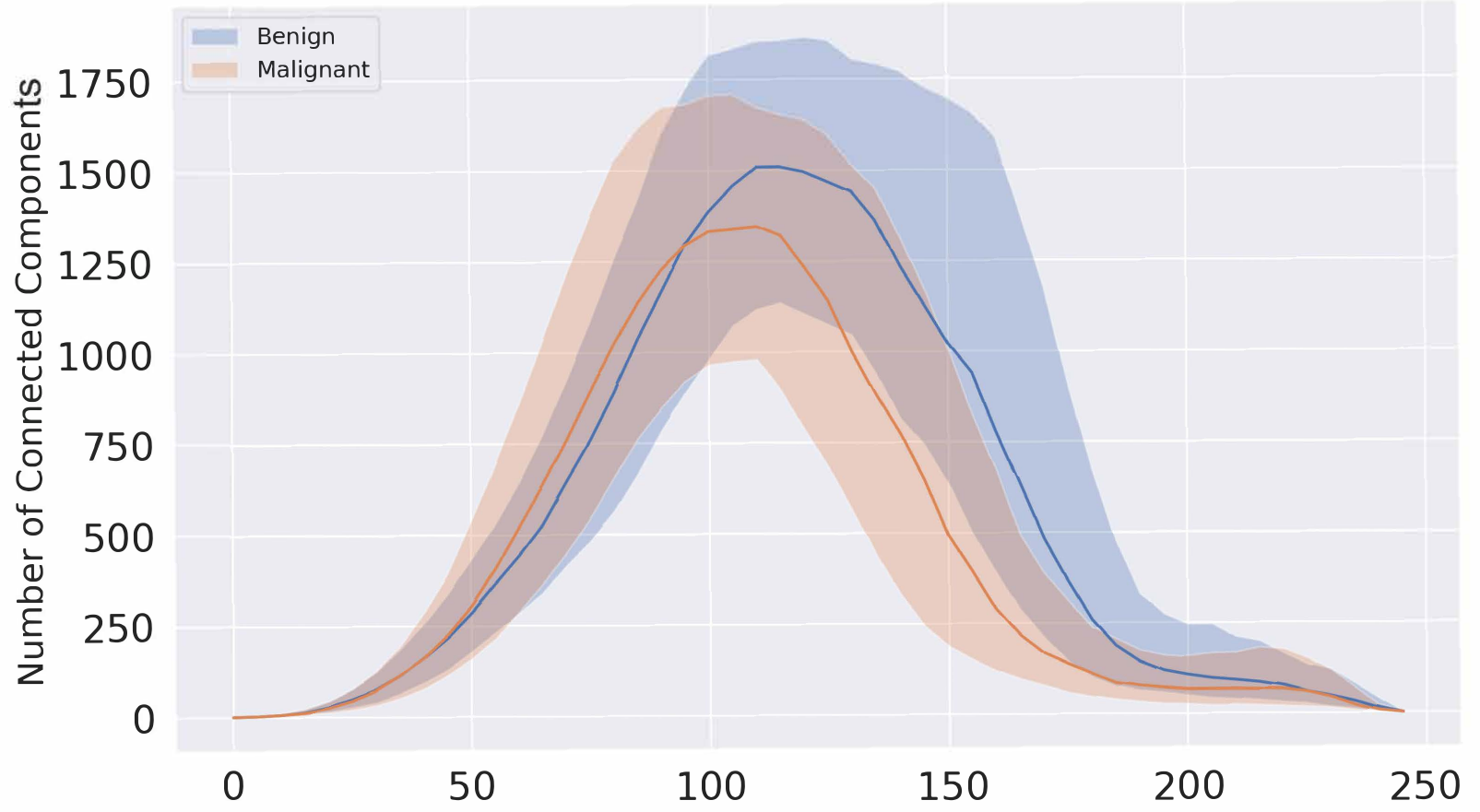}
         \caption{Betti-0 (Gray)}
         \label{fig:Betti-0 (gray)}
     \end{subfigure}
      \hfill
     % \begin{subfigure}[b]{0.32\textwidth}
     %     \centering
     %     \includegraphics[width=\textwidth]{figures/B0_red_40x_BreakHis.pdf}
     %     \caption{Betti-0 (Red)}
     %     \label{fig:Betti-0 (red)}
     % \end{subfigure}
     %  \hfill
     \begin{subfigure}[b]{0.32\textwidth}
         \centering
         \includegraphics[width=\textwidth]{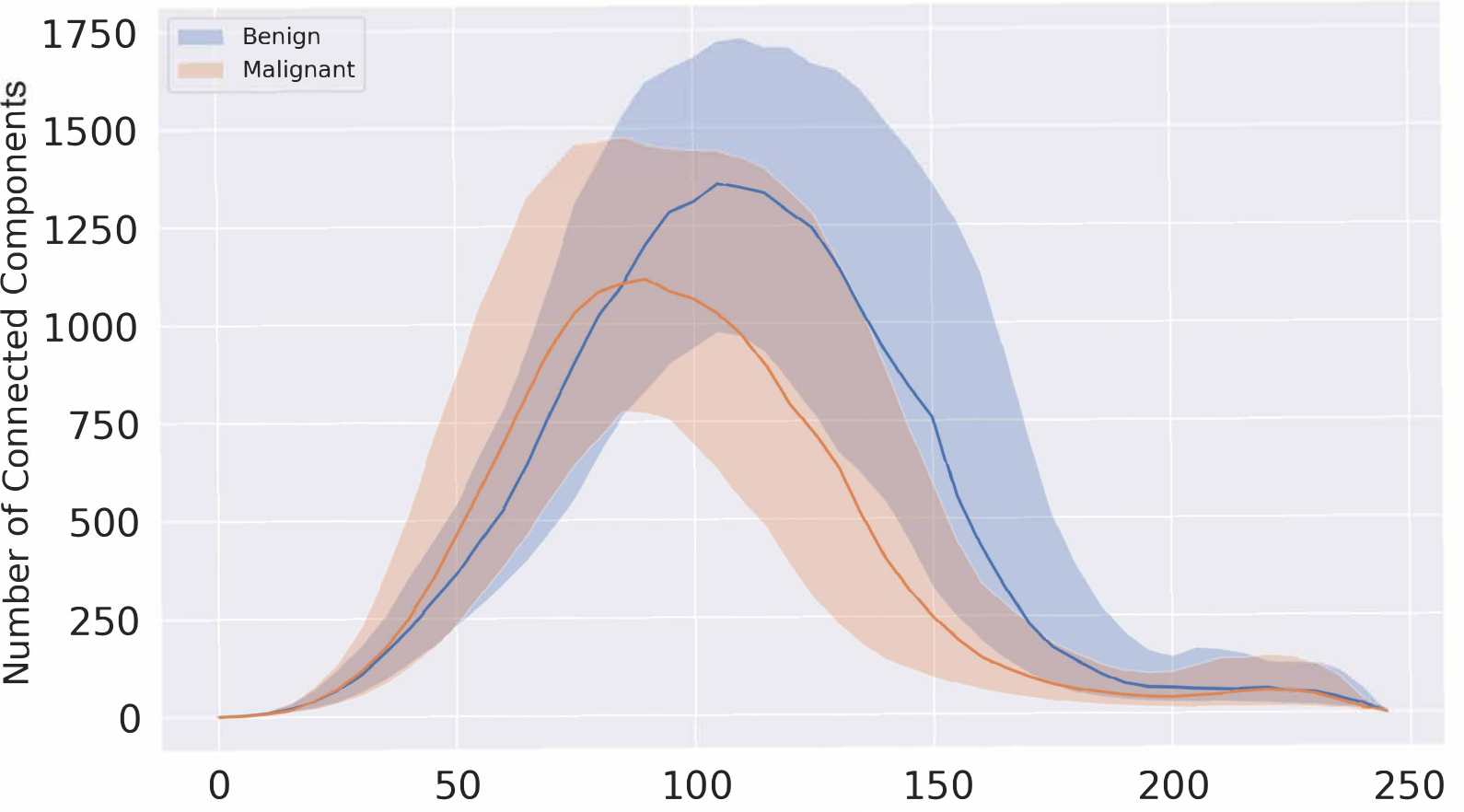}
         \caption{Betti-0 (Green)}
         \label{fig:Betti-0 (green)}
     \end{subfigure}
     % \hfill
     % \begin{subfigure}[b]{0.32\textwidth}
     %     \centering
     %     \includegraphics[width=\textwidth]{figures/B0_blue_40x_BreakHis.pdf}
     %     \caption{Betti-0 (Blue)}
     %     \label{fig:Betti-0 (blue)}
     % \end{subfigure}
     %  \hfill
     % \begin{subfigure}[b]{0.32\textwidth}
     %     \centering
     %     \includegraphics[width=\textwidth]{figures/B1_gray_40x_BreakHis.pdf}
     %     \caption{Betti-1 (Gray)}
     %     \label{fig:Betti-1 (gray)}
     % \end{subfigure}
     % \hfill
     % \begin{subfigure}[b]{0.32\textwidth}
     %     \centering
     %     \includegraphics[width=\textwidth]{figures/B1_red_40x_BreakHis.pdf}
     %     \caption{Betti-1 (Red)}
     %     \label{fig:Betti-1 (red)}
     % \end{subfigure}
      \hfill
     \begin{subfigure}[b]{0.32\textwidth}
         \centering
         \includegraphics[width=\textwidth]{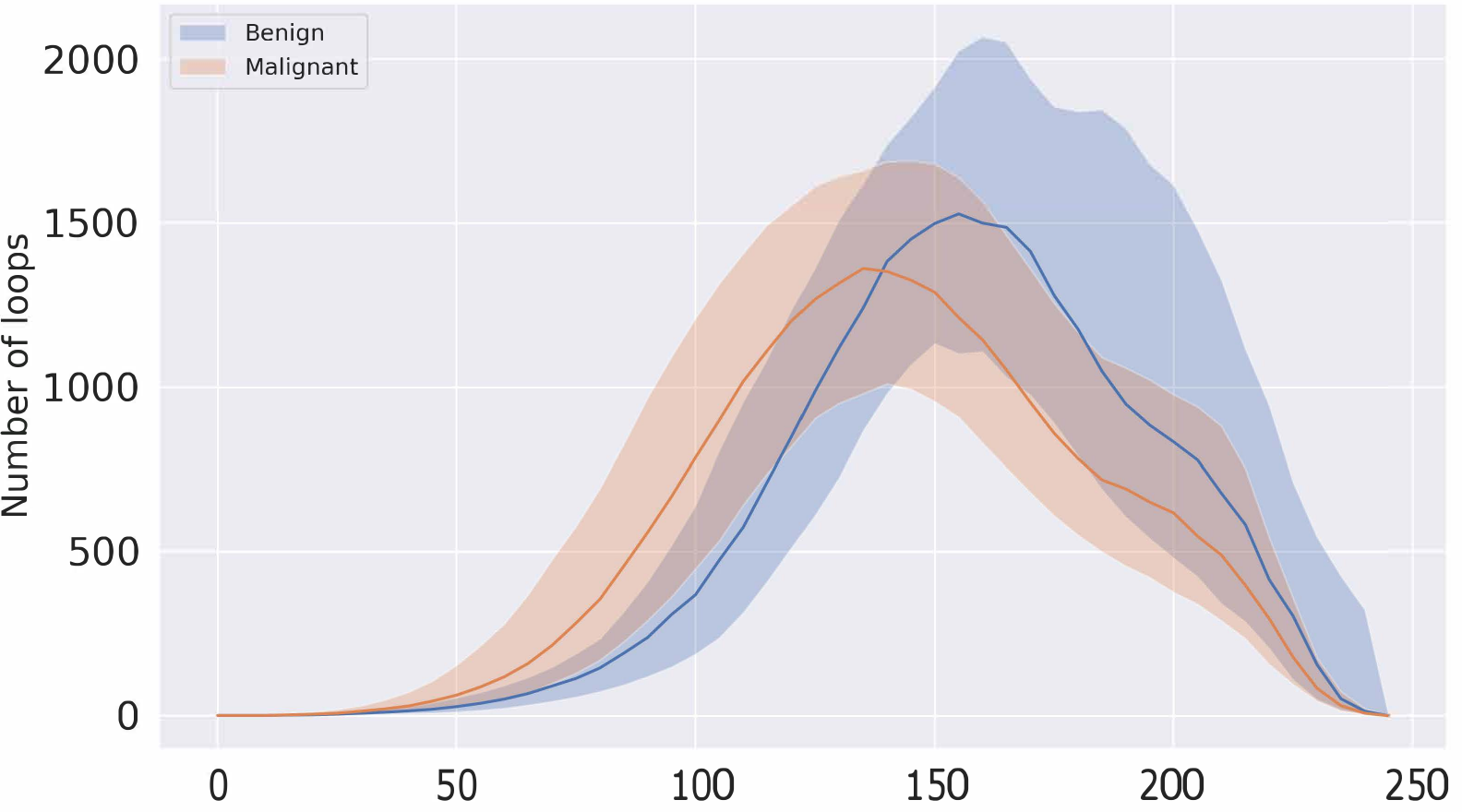}
         \caption{Betti-1 (Green)}
         \label{fig:Betti-1 (green)}
     \end{subfigure}
     % \hfill
     % \begin{subfigure}[b]{0.32\textwidth}
     %     \centering
     %     \includegraphics[width=\textwidth]{figures/B1_blue_40x_BreakHis.pdf}
     %     \caption{Betti-1 (Blue)}
     %     \label{fig:Betti-1 (blue)}
     % \end{subfigure}     
        \caption{Median curves and $40\%$ confidence bands of our topological feature vectors  for each class for BREAKHIS dataset. \label{fig:betti-curves3}}
\end{figure}

In \Cref{fig:betti-curves1,fig:betti-curves3}, we depict our topological vectors and their confidence bands for each class. Using Betti curves for vectorization makes these figures highly interpretable. Specifically, the $x$-axis represents color values $t$ in the range [0, 255]. In Betti-0 curves, the $y$-axis values $\beta_0(t)$ represent the count of connected components in the binary image $\X_t$ (see \Cref{sec:PH_background}). For Betti-1 curves, $\beta_1(t)$ represents the count of holes in the binary image $\X_t$. 

For example, in \Cref{fig:Betti-0-gray (cc-lgsc)}, we observe that between grayscale values $100 < t < 120$, the number of connected components in the binary images $\X_t$ of the CC class is nearly triple that of the LGSC class. This indicates that the binary images of the CC class are highly dispersed with numerous disconnected components compared to the LGSC class. Similarly, in \Cref{fig:Betti-0-gray (hgsc vs. mc)}, between grayscale values $120 < t < 150$, the number of connected components in the binary images $\X_t$ of the HGSC class is almost double that of the MC class.

% \clearpage

%  \pagenumbering{gobble} 

%  \textbf{\Large Changes After Review}\\

%  The original paper was 10 page main text, 2 page references and 1 page appendix. However, in order to proceedings guideline, we moved some of the parts to appendix, and remove some figures from appendix.
% \begin{enumerate}
    
% \item In the experiments section, \textit{hyperparameters, computational complexity and runtime,} and \textit{performance metrics} parts to the Appendix A.\\

%  \item We moved \textit{Visualization and Interpretability} part given after results section to Appendix B.\\
 
%  \item Because of page limitation, we removed two figures from Appendix on visualization of Betti curves.

% \end{enumerate} 

\end{document}